\documentclass{article}

\usepackage[final]{neurips_2019}

\usepackage[utf8]{inputenc} 
\usepackage[T1]{fontenc}    
\usepackage{hyperref}       
\usepackage{url}            
\usepackage{booktabs}       
\usepackage{amsfonts}       
\usepackage{nicefrac}       
\usepackage{microtype}      
\usepackage[countmax]{subfloat}
\usepackage{graphicx}
\usepackage{graphics}
\usepackage{subcaption}
\usepackage{amsmath}
\usepackage{float}
\title{Texture Bias Of CNNs Limits Few-Shot Classification Performance}

\author{%
  Sam Ringer\thanks{Denotes equal contribution.} \qquad Will Williams\footnotemark[1] \qquad \quad  Tom Ash \qquad Remi Francis \qquad David MacLeod \\
  \\
  Speechmatics \\
  Cambridge, UK
  \\
  \\
  \texttt{\{samr,willw\}@speechmatics.com}
}

\begin{document}

\maketitle

\begin{abstract}
Accurate image classification given small amounts of labelled data (few-shot classification) remains an open problem in computer vision. In this work we examine how the known texture bias of Convolutional Neural Networks (CNNs) affects few-shot classification performance. Although texture bias can help in standard image classification, in this work we show it significantly harms few-shot classification performance. After correcting this bias we demonstrate state-of-the-art performance on the competitive \textit{mini}ImageNet task using a method far simpler than the current best performing few-shot learning approaches.

\end{abstract}

\section{Introduction}
The ability of neural networks to perform image classification has increased dramatically in recent years \cite{AlexNet, ResNet, tan2019efficientnet}. However, much of this improvement has relied on using large amounts of labelled data, with successful classification of a given class often requiring thousands of labelled images for training. This is in contrast to the ability of humans to recognize new classes using only one or two labelled examples. The goal of \textit{few-shot classification} is to bridge this gap such that machines can generalize their classification ability to unseen classes using very small amounts of labelled data.

\cite{texturebias} shows that Convolutional Neural Networks (CNNs) show a greater bias towards learning texture-based features than humans. However, they also demonstrate that this bias actually improves classification performance in the standard classification setting, where the classes at test time are drawn from the same distribution as those seen at train time.

At the time of writing there has been no investigation as to how the texture bias of CNNs affects classification performance when a distributional shift in classes is experienced between train and test time, such as that seen in the few-shot learning setting. This work performs this investigation and demonstrates that texture bias significantly decreases performance under distributional shift and correcting for this bias leads to large improvements in few-shot classification accuracy. We focus particularly on how the data \textit{itself} can be manipulated or exploited to aid the learning process.

\section{Related Work}
\subsection{Few-Shot Learning}
A common methodology for evaluating few-shot image classification approaches is to pre-train a model on a corpus of labelled images and then test the model's classification ability on unseen classes, given a small amount of labelled data from said unseen classes. The labelled data from the unseen classes is typically termed the \textit{support} set and the images on which classification is tested is termed the \textit{query} set. A wide range of approaches based on this methodology have been developed \cite{maml, matching, leo, metaoptnet}. 

One distinction between such approaches is that of \textit{parametric} versus \textit{non-parametric} methods. Parametric methods pre-train a model and, when presented with the support set, will \textit{fine-tune} the parameters of the pre-trained model to improve performance on the query set. One such parametric approach is model-agnostic meta-learning (MAML) \cite{maml}, which uses second-order gradients to learn an initialization that can be fine-tuned on a given support set. The resulting model can then perform classification on a corresponding query set. At the time of writing, the best performing parametric approach is classifier synthesis learning (CASTLE) \cite{castle}, which synthesizes few-shot classifiers based on a shared neural dictionary across classes, and then combines these synthesized classifiers with standard 'many-shot' classifiers.

Non-parametric methods perform no fine-tuning when given the support set. Instead, they learn an embedding function and an associated metric space over which classification of new images can be performed. Such approaches include prototypical networks \cite{prototypical} and matching networks \cite{matching}. \cite{prototypical} learn an embedding function that maps images to points in a latent space. For a support set, each class `prototype' is represented by the mean embedding of the examples in the support set. The query set is then classified according to the prototype that minimizes euclidean distance to the embedding of each query image. Non-parametric approaches have shown marginally inferior performance than parametric approaches. However, they are far simpler in their implementation at both at pre-training and test time.

\subsection{Texture Bias Of CNNs}
Outside of the few-shot learning field, there has been great progress in the interrogation of the features learned by CNNs when performing classification. Until recently, it was widely believed that CNNs were able to recognize objects through learning increasingly complicated spatial features \cite{AlexNet}. However, \cite{texturebias} show that CNNs trained on the ImageNet dataset show extreme bias towards learning texture-based representations of images over shape-based representations. Furthermore, \cite{oldhumanvision} show that shape is the single most important feature that humans use when classifying images.

\cite{bagnets} train CNNs with constrained receptive fields, effectively limiting the learned features to only low-frequency local features, such as texture. The resulting model achieves high test accuracies on ImageNet. This shows that texture-based features are adequate for good performance for standard image classification, where train and test classes are drawn from the same distribution.

Taken together, these results pose the question: why do CNNs learn to classify based almost entirely on texture where as humans rely mostly on shape? In this work, we consider a hypothesis: although high-frequency local features (such as texture) generalize well within a distribution, global low-frequency features (such as shape) generalize better under distributional shift. If this hypothesis were true, it could help explain why humans are so heavily dependent on shape-based features; it is because they generalize better to new classes than texture-based features.

\section{Methods}
\subsection{Problem Formulation}

In the typical few-shot classification formulation, a \textit{task} consists of using a labelled \textit{support} set $S = \{(\textbf{x}_i, y_i)\}^{|S|} _{i=1}$ to classify an unseen \textit{query} set $Q = \{\textbf{x}_i\}_{i=1} ^{|Q|}$. The support and query sets are sampled from the same set of classes. During the \textit{pre-training} phase, tasks are sampled from a set of tasks $\mathcal{T} ^{\text{pre-train}}$ and at test time the tasks are sampled from $\mathcal{T} ^{\text{test}}$. There is no class overlap between $\mathcal{T} ^{\text{pre-train}}$ and $\mathcal{T} ^{\text{test}}$. A \textit{k}-shot \textit{n}-way task will contain images from \textit{n} different classes and \textit{k} images from each class. In this work we use the same training scheme, model and loss as \cite{prototypical}. 

\subsection{Stylized Pre-training}
\cite{texturebias} are able to remove the texture bias of CNNs by training on Stylized-ImageNet, which removes each image's texture by using AdaIN style transfer \cite{styletransfer}. For this work, we create an analogous dataset: Stylized-\textit{mini}ImageNet.

 Our pre-training tasks are sampled from Stylized-\textit{mini}ImageNet with probability $p$ and from \textit{mini}ImageNet with probability $1-p$. By sampling from Stylized-\textit{mini}ImageNet with a given probability, we can control the extent to which our model can learn texture-based features as opposed to shape-based features. At test time all tasks are sampled from the withheld classes of standard \textit{mini}ImageNet.

\begin{figure}[h]
 \begin{center}

\begin{subfigure}{0.22\textwidth}
\includegraphics[height=2.4cm]{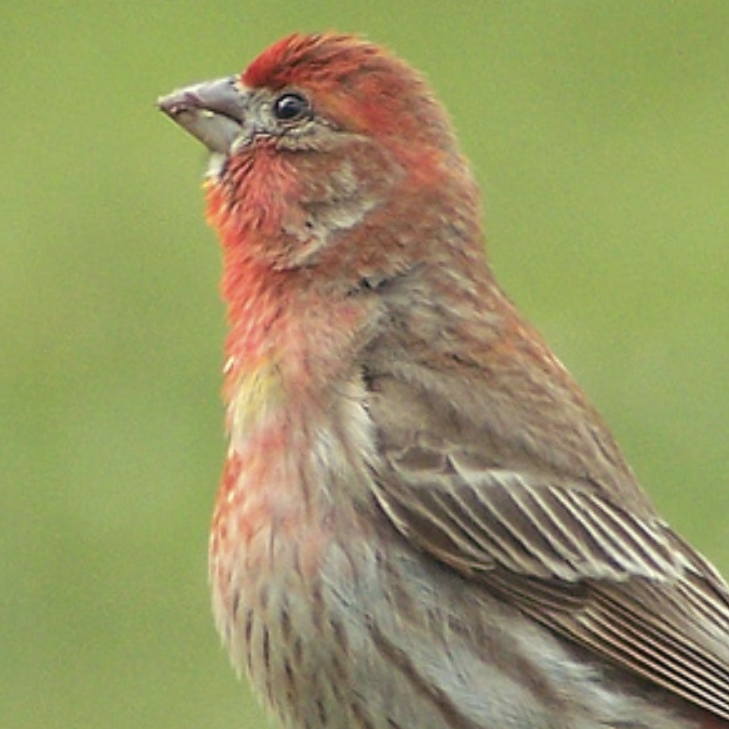} 
\end{subfigure}
\begin{subfigure}{0.22\textwidth}
\includegraphics[height=2.4cm]{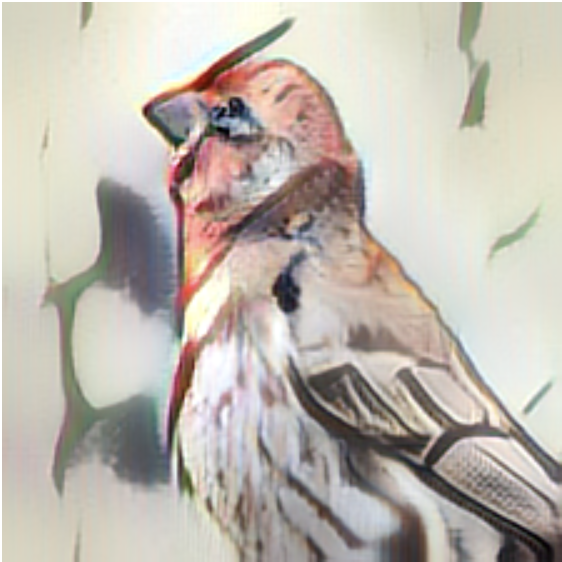}
\end{subfigure}
\begin{subfigure}{0.22\textwidth}
\includegraphics[height=2.4cm]{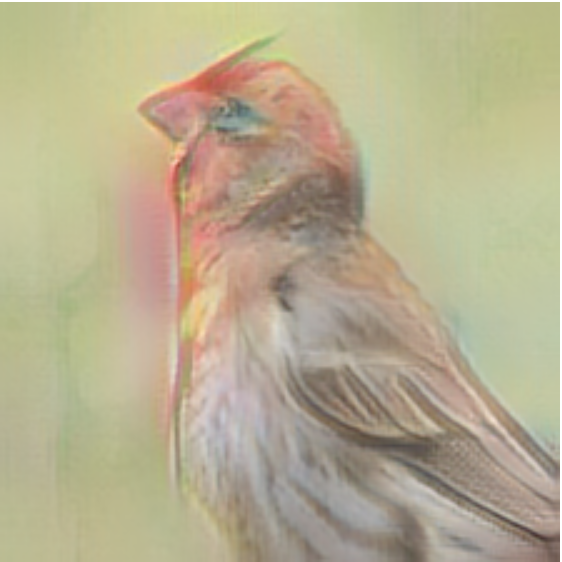} 
\end{subfigure}
\begin{subfigure}{0.22\textwidth}
\includegraphics[height=2.4cm]{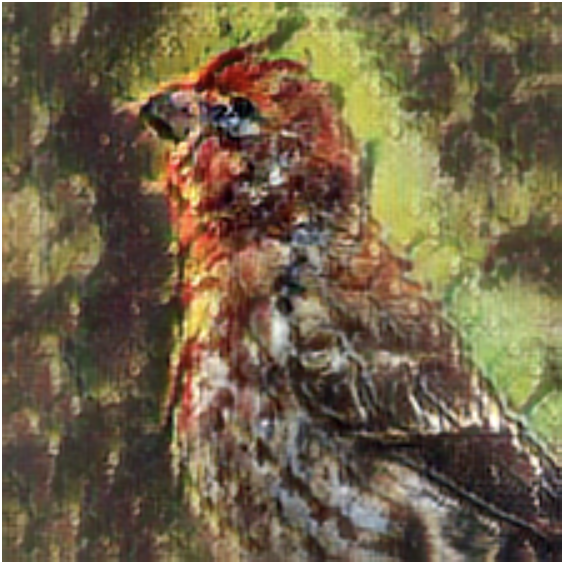}
\end{subfigure}
 
\caption{A sample from Stylized-\textit{mini}ImageNet. The left-most image is the unstylized image with the remaining three being produced using different stylization kernels.}
\label{fig:image2}
 \end{center}
\end{figure}

\subsection{Support \& Query Data Augmentation}
At test-time, as the \textit{k}-shot of a task increases, the problem tends from few-shot to standard many-shot classification and test accuracy increases dramatically. In light of this, we `artificially' increase \textit{k}-shot by performing data-augmentation on the support set. Each image in the support set is augmented $N_{Sa} = 32$ times. Our intention is to investigate the efficacy of emulating the high \textit{k} setting using augmented examples from the support set. The prototype of class $k$, $\textbf{c}_k$, is then given by:
\begin{equation}
    \textbf{c}_k = \frac{1}{|S_k| + N_{Sa} \times |S_k|}\sum_{(x_i,y_i) \in S_k} \left( f_\theta(\textbf{x}_i) + \sum_{i=1}^{N_{Sa}} f_\theta(f_a(\textbf{x}_i))\right)
\end{equation}

$S_k$ is the support set for class $k$. $f_a$ is a randomly sampled data augmentation function. $f_\theta$ is the learned embedding function, in our case a prototypical network \cite{prototypical}.

For the query set, we also augment each image $N_{Qa} = 32$ times. Each of these augmented images is then passed through the embedding function and the estimated probability for a given original query image, \textbf{x}, belonging to class $k$ is given by equations \ref{eq: dist} and \ref{eq: query aug}.

\begin{equation}
\label{eq: dist}
    p_\theta(y=k|\textbf{x}) = \frac{\text{exp}(-d(\textbf{x}, \textbf{c}_k))}{\sum_{k'}\text{exp}(-d(\textbf{x}, \textbf{c}_{k'}))}
\end{equation}

\begin{equation}
\label{eq: query aug}
    d(\textbf{x}, \textbf{c}_k) = \frac{1}{1+N_{Qa}}\left(||f_\theta(\textbf{x}) - \textbf{c}_k||_2 + \sum_{i=1}^{N_{Qa}} || f_\theta(f_a(\textbf{x})) - \textbf{c}_k||_2\right)
\end{equation}

\section{Experiments}

\begin{table}[h]
\smallskip
\caption{\label{tab:main-results}\textbf{Comparison to prior work on 5-shot 5-way \textit{mini}ImageNet.} Conv-x denotes a 4-layer CNN with x filters in each layer.}
\smallskip
\begin{center}
\begin{tabular}{ c c c }
 \toprule
 \textbf{model} & \textbf{backbone} & \textbf{Test Accuracy} \\
 \midrule
 Matching Networks \cite{matching} & Conv-64 & 55.3 $\pm$ 0.7 \\
 MAML \cite{maml}& Conv-32 &  63 $\pm$ 1 \\
 TADAM \cite{tadam} & ResNet-12 & 76.7 $\pm$ 0.3 \\
 ProtoNet (Baseline) \cite{prototypical}& ResNet-12 & 76.8 $\pm$ 0.3 \\ 
 LEO \cite{leo} & WRN-28-10 & 77.6 $\pm$ 0.1 \\  
 MetaOptNet \cite{metaoptnet} & ResNet-12 & 78.6 $\pm$ 0.5\\
 CASTLE \cite{castle} & ResNet-12 & 79.52 $\pm$ 0.02\\
 \midrule
 \textbf{Ours} & ResNet-12 & 80.4 $\pm$ 0.3
\end{tabular}
\end{center}
\end{table}

Table \ref{tab:main-results} shows the performance of our training scheme when testing on \textit{mini}ImageNet for the 5-shot 5-way task. Our method is entirely non-parametric and far simpler to implement at both pre-train and test time than many of the other few-shot classification methods.

\cite{texturebias} show that training on a combination of unstylized and stylized data leads to a small drop in classification accuracy. This is because when the training and testing data are drawn from the same distribution (\textit{i.e.} the same classes) the inherent texture bias of CNNs can actually aid performance.

\begin{figure}[h]
 \begin{center}
\includegraphics[height=5cm]{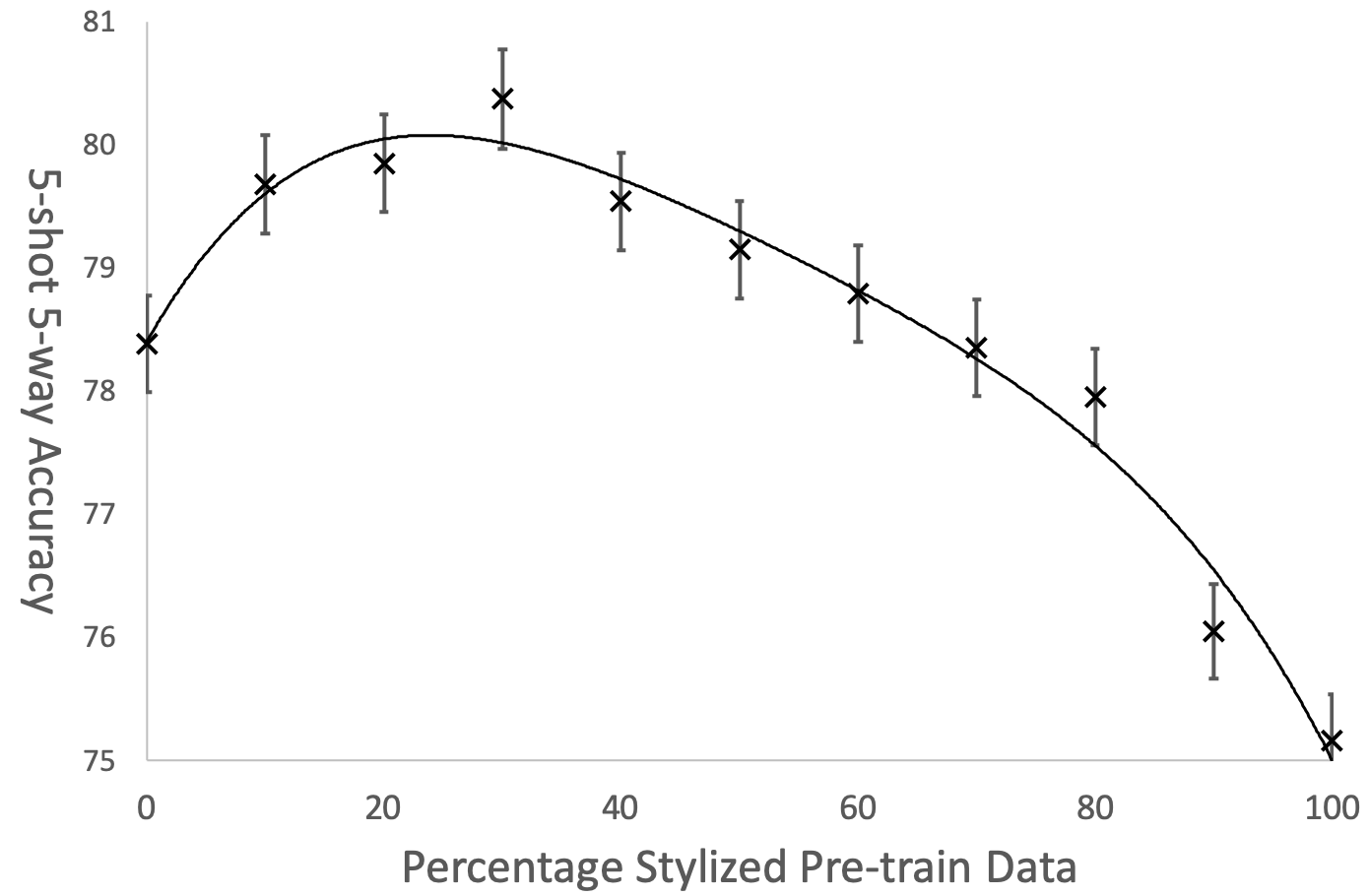}
\caption{5-shot 5-way \textit{mini}ImageNet test accuracy versus pre-training data composition when using test-time augmentation.}
\label{fig:image3}
 \end{center}
\end{figure}

However, in the few-shot learning setting, testing data is drawn from a different distribution to the training data. The ablation shown in Table \ref{tab:ablation} shows that pre-training on a combination of unstylized and stylized data actually \textit{increases} performance at test time, even though the testing data is entirely unstylized. This suggests that the texture bias of CNNs does adversely impact performance under distributional shift.
\begin{table}[h]
\smallskip
\caption{\label{tab:ablation}\textbf{Ablation study.} The effects of pre-training on un-stylized and stylized data, as well as the effects of different forms of test-time augmentation.}
\smallskip
\begin{center}
\begin{tabular}{c c c c c c}
 \toprule
 \textbf{Unstylized Data} & \textbf{Stylized Data} & \textbf{Support TTA} & \textbf{Query TTA} & \textbf{Test Accuracy} \\
 \midrule
 \checkmark & & & &  76.8 $\pm$ 0.3\\ 
  \midrule
 & \checkmark & & & 72.9 $\pm$ 0.3\\ 
  \midrule
 \checkmark & \checkmark &  & &  78.8 $\pm$ 0.3 \\
  \midrule
 \checkmark & \checkmark & \checkmark & & 79.2 $\pm$ 0.3 \\
  \midrule
 \checkmark & \checkmark & \checkmark & \checkmark & 80.4 $\pm$ 0.3
\end{tabular}
\end{center}
\end{table}
Figure \ref{fig:image3} shows that as the proportion of Stylized-\textit{mini}ImageNet pre-training data increases from 0 to 0.3, the accuracy increases as the resulting model is less biased towards texture-based features. However, as the proportion of Stylized-\textit{mini}ImageNet increases from 0.3, the accuracy begins to decrease again as the final model is biased too heavily towards shape-based features.
\section{Conclusion}
It has previously been demonstrated that CNNs are biased towards learning texture-based features over the shape-based features used in human vision. Although this bias aids classification performance when training and testing classes are drawn from the same distribution (standard image classification), this work shows that the texture bias of CNNs significantly decreases classification performance in the few-shot learning setting, where distributional shift is experienced. Correcting for this bias achieves state-of-the-art performance on \textit{mini}ImageNet, even using only a simple non-parametric method.
\medskip

\small

\bibliographystyle{plain}
\bibliography{refs}

\section*{Appendix}
\subsection*{Experimental Setup}
In this work we use a prototypical network \cite{prototypical} with a standard ResNet12 backbone \cite{ResNet}. For training we use the Adam \cite{adam} optimizer with an initial learning rate of $1 \times 10^{-4}$ with parameters $\beta_1 = 0.9$ and $\beta_2 = 0.99$. We do not initialize from any pre-trained weights and our model is trained for 70,000 steps, with the learning rate halving every 15,000 steps. We perform early stopping based off of validation accuracy. We use a temperature of 32 in the SoftMax function in equation \ref{eq: dist}. We test models over 20,000 sampled few-shot tasks, with the mean accuracy and 95\% confidence intervals being reported above. We use no dropout, weight-decay or label smoothing. 

\subsection*{Stylized-\textit{mini}ImageNet}
For the generation of Stylized-\textit{mini}ImageNet, we began with \textit{mini}ImageNet and used the same stylization procedure as \cite{texturebias}. \cite{texturebias} use a stylization coefficient of $\alpha = 1.0$ on ImageNet. When applied to the smaller images of \textit{mini}ImageNet, this stylization coefficient led to images that were so distorted that even humans were unable to perform successful classification. For this reason, we generated Stylized-\textit{mini}ImageNet with a less aggressive stylization coefficient of $\alpha = 0.4$.

To ensure diversity of styles and true independence of texture and underlying image, we generate 10 stylized images for each original \textit{mini}ImageNet image. The stylization was performed only on the \textit{train} split of \textit{mini}ImageNet as testing and validation were both done on standard \textit{mini}ImageNet.

\subsection*{Data Augmentation}
At train and test time, we apply a standard set of data augmentations. The applied data augmentations are as follows: random horizontal flip, random brightness jitter, random contrast jitter, random saturation jitter and random crop between 70\% and 100\% of original image size. The final image is re-sized to be of size 84x84 pixels.

\end{document}